\documentclass{article}
\usepackage{spconf,amsmath,graphicx}
\usepackage{amsfonts}
\usepackage[ruled,linesnumbered]{algorithm2e}
\usepackage{multirow}

\DeclareMathOperator{\rank}{rank}
\DeclareMathOperator{\Null}{Null}

\title{Learning a low-rank feature representation: Achieving Better Trade-off between Stability and Plasticity in Continual Learning}

\name{Zhenrong Liu$^{\star,\parallel}$, Yang Li$^{\dagger}$, Yi Gong$^{\star}$, and Yik-Chung Wu$^{\parallel}$}
\address{\small $^{\star}$ Department of Electrical and Electronic Engineering, Southern University of Science and Technology, Shenzhen, China\\
 \small $^{\parallel}$ Department of Electrical and Electronic Engineering, The University of Hong Kong, Hong Kong\\
 \small $^{\dagger}$ Shenzhen Research Institute of Big Data, Shenzhen, China}
\newtheorem{theorem}{Theorem}
\newtheorem{lemma}[theorem]{Condition}

\begin{document}

\onecolumn

{\fontsize{14}{16}\selectfont IEEE Copyright Notice}

\

\copyright~2023 IEEE. Personal use of this material is permitted. Permission from IEEE must be obtained for all
other uses, in any current or future media, including reprinting/republishing this material for advertising
or promotional purposes, creating new collective works for resale or redistribution to servers or lists, or
reuse of any copyrighted component of this work in other works.

\newpage
\twocolumn

\maketitle
\begin{abstract}
In continual learning, networks confront a trade-off between stability and plasticity when trained on a sequence of tasks. To bolster plasticity without sacrificing stability, we propose a novel training algorithm called LRFR. This approach optimizes network parameters in the null space of the past tasks' feature representation matrix to guarantee the stability. Concurrently, we judiciously select only a subset of neurons in each layer of the network while training individual tasks to learn the past tasks' feature representation matrix in low-rank. This increases the null space dimension when designing network parameters for subsequent tasks, thereby enhancing the plasticity. Using CIFAR-100 and TinyImageNet as benchmark datasets for continual learning, the proposed approach consistently outperforms state-of-the-art methods.
\end{abstract}
\section{Introduction}
\label{sec:intro}
Continual learning aims to equip deep neural networks (DNNs) with the capacity to acquire new knowledge while retaining old information, similar to human learning. However, DNNs can suffer from catastrophic forgetting, leading to performance degradation on past tasks when learning new ones sequentially. This challenge highlights the need for effective solutions in continual learning algorithms.

This challenge is tied to the stability-plasticity dilemma \cite{mirzadeh2020understanding}, where networks need both the plasticity to learn new information and the stability to retain previous knowledge. To address this issue, existing continual learning methods fall into the following four categories: architecture-based approaches expand networks \cite{jerfel2019reconciling,mallya2018packnet}; replay-based methods blend old and new data \cite{isele2018selective,rolnick2019experience}; regularization-based methods penalize past task updates, e.g., EWC \cite{kirkpatrick2017overcoming}; and algorithm-based methods modify the parameter update rules \cite{lopez2017gradient,saha2021gradient,wang2021training,chaudhry2018efficient}. Different categories of methods can coexist in a continual learning task, and this paper focuses on algorithm-based methods.

In algorithm-based methods, a common strategy is to project parameter gradients into the null space of the past tasks' feature representation matrix, ensuring that the new task training does not affect previous tasks' performance \cite{saha2021gradient,wang2021training}. However, since the trainable parameters are restricted within the null space, the plasticity may be degraded when the null space dimension is small. To address this issue, existing methods \cite{saha2021gradient,wang2021training,kong2022balancing} approximate the past tasks' feature representation matrix with low-rank techniques before gradient projection, yet this compromises the stability since the parameter updates are pushed outside the actual representation matrix's null space.

To address this stability-plasticity dilemma, we introduce a novel algorithm that bolsters plasticity while upholding stability by learning a low-rank feature representation (LRFR) \footnote{https://github.com/Dacaidi/LRFR}. By establishing a link between the rank of the past tasks' feature representation matrix and its null space dimension, we reveal that a low-rank representation matrix enhances the network's plasticity. To achieve this, we judiciously select a subset of neurons in each network layer to reduce the rank of the representation matrix. In summary, our contributions can be distilled into two main aspects:

1. We demonstrate that a lower rank of the past tasks' feature representation matrix enhances the plasticity in algorithm-based gradient methods for continual learning.

2. To enhance the plasticity, we strive to learn a low-rank past tasks' feature representation matrix. Diverging from existing methods that utilize low-rank approximations to expand the gradient projection space outside the null space of the past tasks’ feature representation matrix, the proposed approach increases the null space dimension of the actual feature representation matrix and hence maintains the network's stability.

\section{Preliminaries}
\label{sec:format}
\subsection{Settings and Notations}
We consider a sequence of tasks where $T$ denotes the total number of tasks. Let $t=1, 2, \ldots T$ and  $\mathcal{D}_t=\left\{\mathcal{X}_t, \mathcal{Y}_t\right\}$ be the dataset of task $t$, where $\mathcal{X}_t=\left\{\boldsymbol{x}_{i, t}\right\}_{i=1}^{n_t}$ and $\mathcal{Y}_t=\{\boldsymbol{y}_{i, t}\}_{i=1}^{n_t}$ are the corresponding input and label sets with $n_t$ denoting the number of samples. In continual learning, the model is trained on these datasets sequentially. 

Consider an $L$ layer neural network where each layer computes the following function for task $t$:
\begin{equation}
    \boldsymbol{x}_{i, t}^{l+1} = \sigma\left(\text{BN}\left(\left(\boldsymbol{W}^{l}_{t}\right)^{\mathrm{T}}\boldsymbol{x}_{i, t}^l\right)\right).
\end{equation}
Here, $l=1, 2, \ldots L$, $\sigma(\cdot)$ is a non-linear function, and $\text{BN}(\cdot)$ is the batch normalization layer. At the first layer, $\boldsymbol{x}_{i, t}^1=\boldsymbol{x}_{i, t}$ represents the raw input data. In the subsequent layers, we define $\boldsymbol{x}_{i, t}^l$ as the representation of the input $\boldsymbol{x}_{i, t}$ at layer $l$. The set of parameters of the network is defined by $\mathcal{W}_t=\{\boldsymbol{W}_t^l\}_{l=1}^L$. Here, $\boldsymbol{W}^{l}_{t} \in \mathbb{R}^{a_l \times b_l}$ where $a_l$ and $b_l$ denotes the input and output size of layer $l$, respectively.
Let $\Delta \boldsymbol{W}_{t,s}^l$ represent the parameter update during the training task $t$ at layer $l$ of the network, occurring at training step $s$. Consequently, $\Delta\boldsymbol{W}_t^l$ is the accumulated sum of all such updates across steps. 
Furthermore, the parameter updated after task $t$ is denoted as $\Delta\mathcal{W}_t=\{\Delta\boldsymbol{W}_t^l\}_{l=1}^L=\{(\boldsymbol{W}_t^l-\boldsymbol{W}_{t-1}^l)\}_{l=1}^L$. 
We denote the representation matrix of the raw input $\mathcal{X}_t$ at layer $l$ as $\boldsymbol{F}_{t}^{l}=[\boldsymbol{x}_{1, t}^l, \boldsymbol{x}_{2, t}^l, \cdots, \boldsymbol{x}_{n_t, t}^l] \in \mathbb{R}^{a_{l} \times n_t}$.

\subsection{Null Space}
Suppose $\Delta \boldsymbol{W}_{t,s}^l$ lies in the null space of the past tasks' feature representation matrix. In that case, the stability can be guaranteed, which is illustrated by the following condition.
\begin{lemma}
\label{l1}
(Stability) \cite{wang2021training}. Let $\Bar{n}_{t-1}$ be the the total number of seen data before task $t$, $\tilde{\boldsymbol{F}}^l_{t-1}=\left[\boldsymbol{F}_{1}^l, \boldsymbol{F}_{2}^l, \ldots, \boldsymbol{F}_{t-1}^l\right]\in \mathbb{R}^{a_{l} \times \Bar{n}_{t-1}}$, $f\left(\mathcal{W}_p; \mathcal{X}_p\right)$ be the output of the L-layer network with parameters $\mathcal{W}_p$ when the network is fed with data $\mathcal{X}_p$. If at each training step of task $t$, $\Delta \boldsymbol{W}_{t,s}^l$ lies in the null space of the past tasks' feature representation matrix $\boldsymbol{\Bar{F}}_{t-1}^l=\frac{1}{\Bar{n}_{t-1}}\tilde{\boldsymbol{F}}^l_{t-1}\left(\tilde{\boldsymbol{F}}^l_{t-1}\right)^{\mathrm{T}}$, i.e.,
\begin{equation}
\label{2}
\boldsymbol{\Bar{F}}_{t-1}^l \Delta \boldsymbol{W}_{t,s}^l=\mathbf{0}, \quad \forall l=1, 2, \ldots, L, 
\end{equation}
then we have $f\left(\mathcal{W}_t; \mathcal{X}_p\right)=f\left(\mathcal{W}_{t-1}; \mathcal{X}_p\right)$ for all $p \in\{1, 2, \ldots, t-1\}$ if the batch normalization parameters are fixed.
\end{lemma}

According to Condition \ref{l1}, if the parameters are updated in the null space of $\boldsymbol{\Bar{F}}_{t-1}^l$, then the training loss of the past tasks will be retained, and thus the forgetting can be avoided.
\begin{lemma}
\label{l2}
    (Plasticity) \cite{wang2021training}. Given a network being trained on task $t$, $\boldsymbol{G}_{t, s}^{l}$ represents the gradient direction for layer $l$'s parameters at training step $s$ and $\langle\cdot, \cdot\rangle$ represents the inner product. 
    If $\Delta \boldsymbol{W}_{t,s}^l$ satisfies $\!\langle \Delta \!\boldsymbol{W}_{t, s}^l, \boldsymbol{G}_{t, s}^{l}\rangle\!>\!0$ during the training on task $t$ for all layers, $\Delta \boldsymbol{W}_{t,s}^l$ can reduce the training loss, allowing the network to acquire knowledge from this task.
\end{lemma}

In this paper, we strive to propose a training algorithm to satisfy both the above conditions simultaneously so that it can ensure the network's plasticity while maintaining the stability.

\section{LEARNING A LOW-RANK FEATURE REPRESENTATION}
\label{sec:majhead}
\subsection{Stability and Plasticity Trade-off}
The stability condition requires that each column of $\Delta \boldsymbol{W}_{t,s}^l$ lies within the null space of the past tasks' feature representation matrix $\boldsymbol{\Bar{F}}_{t-1}^l$. Here, we can obtain the dimension of $\boldsymbol{\Bar{F}}_{t-1}^l$'s null space as
\begin{equation}
\label{dof}
    \Null(\boldsymbol{\Bar{F}}_{t-1}^l)=a_l-\rank(\boldsymbol{\Bar{F}}_{t-1}^l).
\end{equation}
Equation (\ref{dof}) holds as $\boldsymbol{\Bar{F}}_{t-1}^l$ is a square matrix, and its rank signifies its range space dimension. 

Equation (\ref{dof}) suggests that a reduced rank of $\boldsymbol{\Bar{F}}_{t-1}^l$ enhances the plasticity. This is because a larger null space dimension allows for the update of $\Delta \boldsymbol{W}_{t,s}^l$ in a higher-dimensional vector space, making it easier to satisfy Condition 2. However, as the sequential training goes on, the rank of $\boldsymbol{\Bar{F}}_{t-1}^l$ grows due to the added data. This in turn reduces the null space dimension and limits the design space of $\Delta \boldsymbol{W}_{t,s}^l$, degrading the plasticity.

To enlarge the design space of $\Delta \boldsymbol{W}_{t,s}^l$, existing approaches commonly relax Condition \ref{l1} by replacing $\boldsymbol{\Bar{F}}_{t-1}^l$ in (\ref{2}) by a low-rank approximate matrix. Techniques like SVD-based principal component selection \cite{saha2021gradient,wang2021training} or rank-constrained Frobenius norm minimization \cite{kong2022balancing} are adopted. However, this impairs the stability as the $\Delta \boldsymbol{W}_{t,s}^l$ will be beyond the null space of $\boldsymbol{\Bar{F}}_{t-1}^l$, causing catastrophic forgetting.

To achieve a better trade-off between stability and plasticity, we learn a low-rank $\boldsymbol{\Bar{F}}_{t-1}^l$ during the continual learning.
Rather than initially acquiring $\boldsymbol{\Bar{F}}_{t-1}^l$ and subsequently employing a low-rank approximation, our approach directly learns $\boldsymbol{\Bar{F}}_{t-1}^l$ in a low-rank manner, so that the plasticity is enhanced without sacrificing the stability.

\subsection{Increasing Plasticity via LRFR}
\label{RUR}
\begin{figure}[htb]
\centering
\includegraphics[width=3.35in]{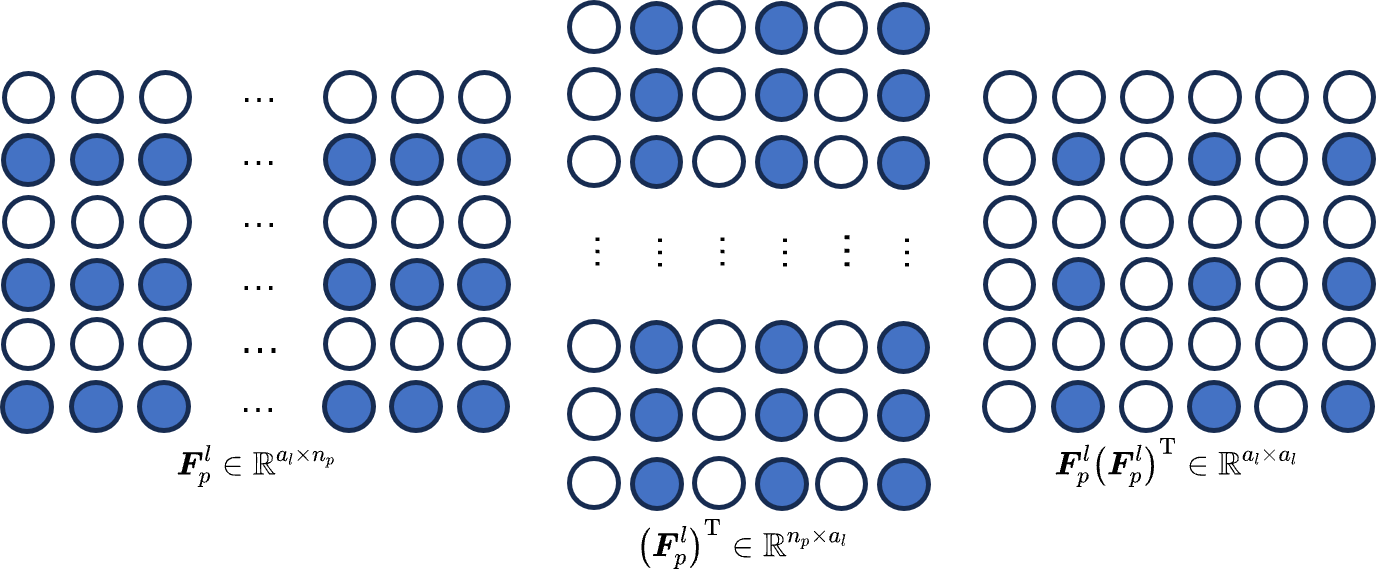}
\caption{The illustration of LRFR. Each circle represents an element within the matrix. Colored circles denote non-zero elements, corresponding to active neurons, while non-colored circles represent zero elements, signifying disabled neurons.}
\label{FRR}
\end{figure}

First, we rewrite $\boldsymbol{\Bar{F}}_{t}^{l}$ in Condition 1 as
\begin{equation}
\label{4}
\boldsymbol{\Bar{F}}_{t}^{l}=\frac{1}{\Bar{n}_{t}}\boldsymbol{\tilde{F}}^{l}_{t}\left(\boldsymbol{\tilde{F}}^{l}_{t}\right)^{_{\mathrm{T}}}
    =\frac{1}{\bar{n}_{t}} \sum\nolimits_{p=1}^{t}\boldsymbol{F}_{p}^{l}\!\left(\boldsymbol{F}_{p}^{l}\right)^{_{\mathrm{T}}}.
\end{equation}
In equation (\ref{4}), reducing the rank of the feature representation matrix $\boldsymbol{F}_{p}^{l}\!\left(\boldsymbol{F}_{p}^{l}\right)^{_{\mathrm{T}}}$ for each $p$ can potentially lead to a lower-rank $\boldsymbol{\Bar{F}}_{t}^{l}$. This is because the upper limit of the rank for the combined matrix $\boldsymbol{\Bar{F}}_{t}^{l}$ is determined by the sum of ranks of the constituent matrices $\boldsymbol{F}_{p}^{l}\!\left(\boldsymbol{F}_{p}^{l}\right)^{_{\mathrm{T}}}$. Moreover, it can be observed that the rank of $\boldsymbol{F}_{p}^{l}\!\left(\boldsymbol{F}_{p}^{l}\right)^{_{\mathrm{T}}}$ is upper bounded by the row rank of $\boldsymbol{F}_{p}^{l}$ since $\boldsymbol{F}_{p}^{l}\in \mathbb{R}^{a_{l} \times n_{p}}$ and ${n}_{p} \gg a_l $. Therefore, in the training process for task $p$, if we can disable a portion of neurons at layer $l$, we can allow the network to learn a low-rank $\boldsymbol{F}_{p}^{l}$, and hence largely decrease the rank of $\boldsymbol{F}_{p}^{l}\!\left(\boldsymbol{F}_{p}^{l}\right)^{_{\mathrm{T}}}$. The low-rank structure of $\boldsymbol{F}_{p}^{l}\!\left(\boldsymbol{F}_{p}^{l}\right)^{_{\mathrm{T}}}$ is shown in Fig \ref{FRR}. It can be seen that with $d_{t,l}$ disabled neurons, we can obtain a low-rank feature representation for the past tasks.

Next, we present how to select the most suitable neurons for disabling. In continual learning, the capacity of a DNN often exceeds the information of a single task, leading to over-parameterization. Leveraging the lottery ticket hypothesis, which indicates sub-networks with fewer neurons can match or surpass the performance of the original larger network \cite{liu2018rethinking, frankle2018the}, we can use this insight to identify disabled neurons. This involves obtaining a sub-network with fewer neurons per layer while maintaining or enhancing performance, revealing critical neurons for the task's performance. Consequently, non-included neurons in this sub-network can be safely disabled during training. This ensures that a low-rank $\boldsymbol{\Bar{F}}_{t}^l$ can be learned after training task $t$ without compromising the performance.

\begin{figure*}[htb]
\centering
\includegraphics[width=4.5in]{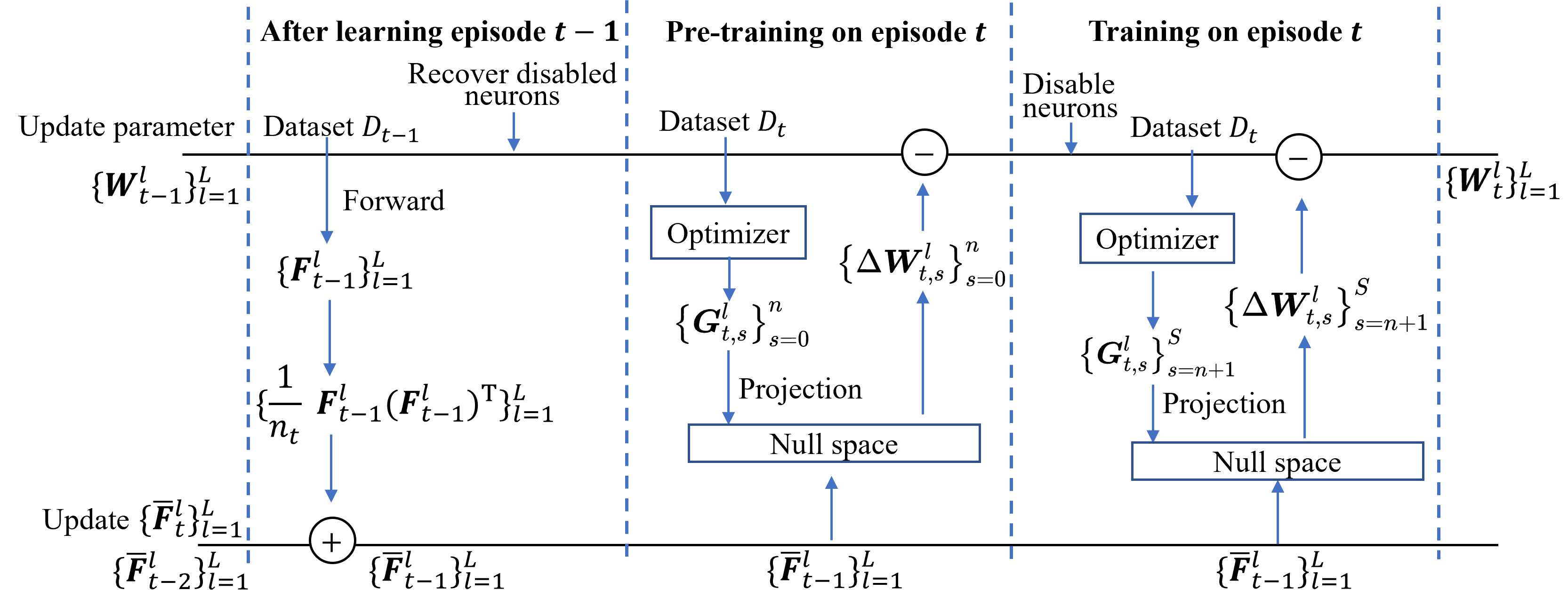}
\caption{Proposed training algorithm pipeline with vertical dashed line separation between stages.}
\label{pipeline}
\end{figure*}

Specifically, we maintain
the layer-wise $\boldsymbol{\Bar{F}}_t^l$, which are incrementally updated using $\boldsymbol{{F}}_t^l$ of the coming task. Given the current task, starting from the network trained on previous tasks, the parameter update is achieved by projecting optimizer-generated updates $\boldsymbol{G}_{t, s}^{l}$ into the null space of $\boldsymbol{\Bar{F}}_{t-1}^l$ layer by layer. The visual representation of this algorithm is shown in Figure \ref{pipeline}. The training process for a given task $t>1$ involves three essential stages. (For the first task $t=1$, only the second and third stages are included when training.)

The initial stage focuses on updating $\boldsymbol{\Bar{F}}_{t-1}^l$. This entails performing forward propagation on the dataset $\mathcal{D}_{t-1}$ to facilitate the update of $\boldsymbol{\Bar{F}}_{t-1}^l$. Applying the proposed LRFR sets this stage apart from existing methods. In our case, the forward propagation does not utilize the entire network; instead, it employs the subnetwork selected from the pretraining stage of the previous task to generate the low-rank $\boldsymbol{F}_{t-1}^{l}$, as will be introduced in the following.

In the pretraining stage for task $t$, we decide which neurons to disable for LRFR in the official training. In this stage, all network neurons are active. Drawing inspiration from \cite{Liu_2017_ICCV}, we add an extra penalty term to the batch normalization layer parameters during network pretraining, leading to the following loss function:
\begin{equation}
\label{sparityloss}
L=\sum_{\mathcal{D}_t} l(f(\mathcal{W}_{t};\mathcal{X}_t), \mathcal{Y}_t)+\mu \sum_{\gamma \in \Gamma} |\gamma|,
\end{equation}
where $l(\cdot)$ is the standard loss function, $\mu$ is the penalty parameter, and $\Gamma$ represents the set of scale parameters in the batch normalization layers. Minimizing (\ref{sparityloss}) during pretraining enforces sparsity in the scale parameters of batch normalization layers. This sparsity structure leverages network architecture and information from $\mathcal{D}_t$, indicating which neurons should be disabled in the subsequent official training stage. Specifically, larger-scale parameters significantly influence layer outputs, contributing more to the next layer's input. Consequently, neurons linked to these significant parameters are selected for sub-network inclusion, while the rest are disabled. Such a designed sub-network is used in the third training stage. Due to dataset disparities, different tasks may have distinct sub-networks in this stage.

In the final stage, we pinpoint the uppermost $k$ percent of neurons possessing the most influential scaling parameters from the pretraining stage. We disable the remaining neurons and conduct official training on dataset $\mathcal{D}_{t}$. This selected sub-network is subsequently employed in the initial stage of task $t+1$, significantly reducing the rank of $\boldsymbol{F}_t^{l}$. This cycle repeats, with the same stages for each subsequent task.
\section{EXPERIMENTAL RESULTS}

\textbf{Datasets}: 
We evaluate our approach on CIFAR-100 and TinyImageNet. In particular, CIFAR-100 is divided into 10 and 20 tasks with non-overlapping classes, i.e., 10 and 20-split-CIFAR-100, respectively. Moreover, TinyImageNet is partitioned into 25 tasks with non-overlapping classes to obtain the more challenging 25-split-TinyImageNet.

\textbf{Implementation details}: 
In PyTorch, we use ResNet-18 as the base network. Tasks share this backbone with separate classifiers and batch normalization parameters \cite{wang2021training}. 
We start with a learning rate of $5\times 10^{-5}$, halving it at epochs 30 and 60 over 80 total epochs. Batch sizes are 32 (CIFAR) and 16 (TinyImageNet). We set the penalty parameter $\mu=0.1$  for neuron selection and disabled $k=50\%$ neurons in the network.

\textbf{Evaluation metrics}:
We adopt evaluation metrics from \cite{lopez2017gradient}: backward transfer (BWT) and average accuracy (ACC). BWT quantifies accuracy decline on previous tasks after learning new ones, indicating forgetting. ACC represents mean accuracy across all seen task test datasets. In cases of comparable ACC, a higher BWT is favored.

\begin{table}
\label{t1}
\small
\centering
\begin{tabular}{l|c|c|c}
\hline
\textbf{Method} & \textbf{10-CIFAR} & \textbf{20-CIFAR} & \textbf{25-Tiny} \\
\cline{2-4}
& \textbf{ACC (BWT)} & \textbf{ACC (BWT)} & \textbf{ACC (BWT)} \\
\hline
\hline
EWC \cite{kirkpatrick2017overcoming}& 70.77 (-2.83) & 71.66 (-3.72) & 52.33 (-6.17) \\
MAS \cite{aljundi2018memory} &66.93(-4.03) & 63.84(-6.29) & 47.96 (-7.04) \\
GEM \cite{lopez2017gradient}& 49.48 (2.77) & 68.89 (-1.2) & N/A \\
A-GEM \cite{chaudhry2018efficient}& 49.57 (-1.13) & 61.91 (-6.88) & 53.32 (-7.68) \\
MEGA \cite{guo2020improved}& 54.17 (-2.19) & 64.98 (-5.13) & 57.12 (-5.90) \\
OWM \cite{zeng2019continual}& 68.89 (-1.88) & 68.47 (-3.37) & 49.98 (-3.64) \\
GPM \cite{saha2021gradient} & $73.66$ (-2.20) & $75.20$ (-7.58) & $58.96$ (-6.96) \\
NSCL\cite{wang2021training}& $73.77$ (-1.6) & $75.95$ (-3.66) & $58.28$ (-6.05) \\
AdNS \cite{kong2022balancing} & 77.21(-2.32) & 77.33(-3.25) & 59.77(-4.58)\\
\hline
LRFR & $\mathbf{81.30}$ (0.11) & $\mathbf{82.95}$ (-1.37) & $\mathbf{62.28}$ (-4.05) \\
\hline
\end{tabular}
\caption{Experimental results on different datasets.}
\end{table}


We present the comparative outcomes of our proposed approach alongside various baseline methods in Table 1, demonstrating the highest ACC and comparable BWT across benchmarks. Compared to regularization-based approaches like EWC and MAS, our method achieves over 10\% higher ACC with less forgetting. In algorithm-based comparisons, on CIFAR-100's 10 and 20-split, our method outperforms all the other baseline methods in ACC and BWT. On 25-Split TinyImageNet, our ACC surpasses all methods, with BWT slightly lower than OWM. These results affirm the proposed LRFR's ability to enhance plasticity while maintaining stability through learned low-rank past tasks' feature representation matrix.
\begin{figure}[htb]
\centering
\includegraphics[width=3.35in]{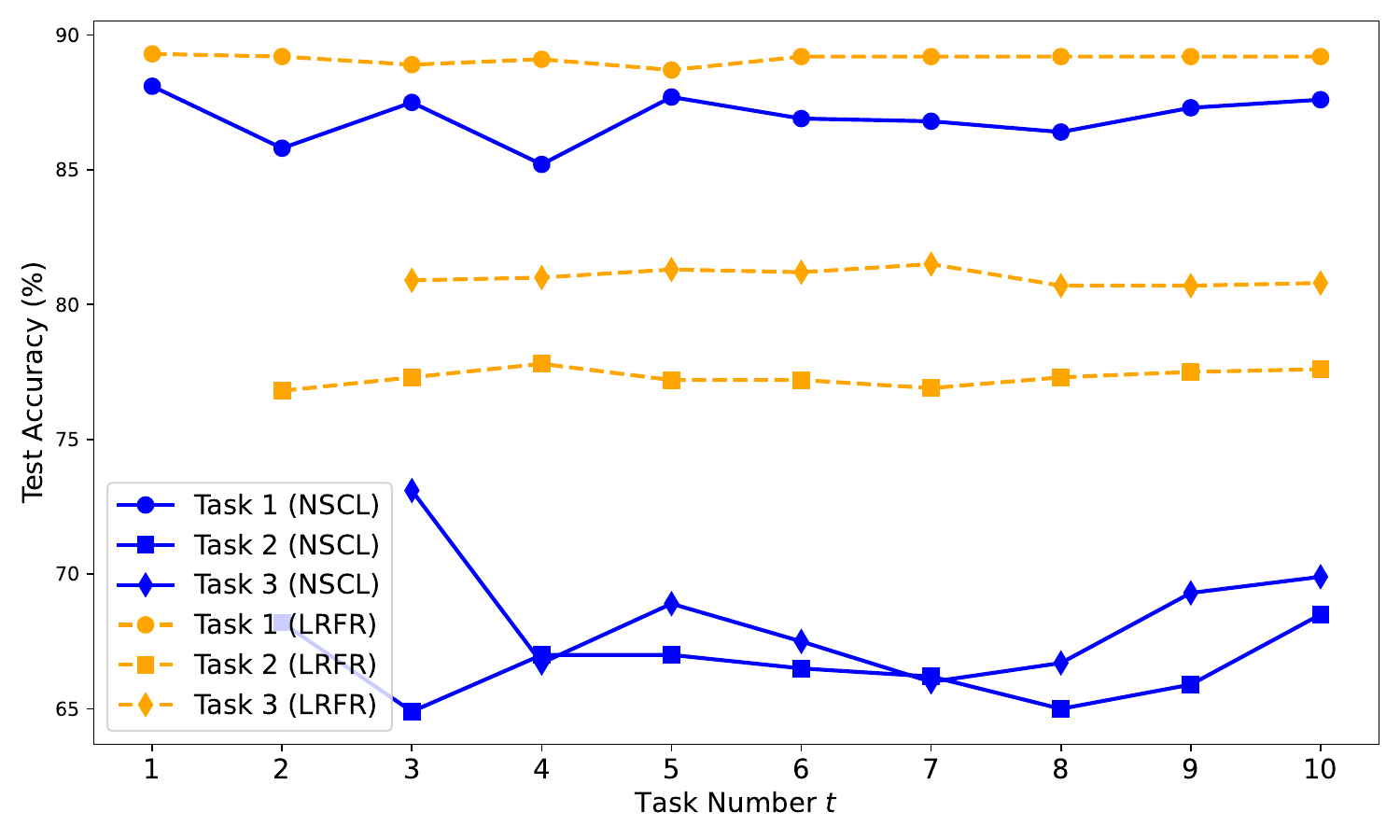}
\caption{The curves of test accuracy on tasks 1, 2, and 3 when the network is trained sequentially.}
\label{taskp}
\end{figure}

Figure \ref{taskp} depicts the test accuracy curves achieved through sequential network training for tasks 1, 2, and 3 on the 10-split CIFAR dataset. We compared LRFR with NSCL since NSCL performed superior to all other open-source codes in Table 1. At $t=1$, it becomes evident that LRFR outperforms NSCL by 1.3\%, emphasizing that disabling neurons does not negatively impact the test accuracy of single tasks. Across ensuing tasks, LRFR consistently exceeds NSCL by 7\%+, showcasing enhanced plasticity. In addition, our approach maintains smoother and superior test accuracy during sequential training, accentuating its better stability and enhanced plasticity over NSCL.

\section{Conclusions}
In this paper, we address the plasticity-stability dilemma for continual learning. We quantitatively illustrate the stability-plasticity relationship in algorithm-based methods and propose the LRFR algorithm. By projecting candidate parameter updates into the null space with higher dimensions, LRFR increases the network plasticity while maintaining stability. Extensive experiments show that the proposed algorithm outperforms the compared methods for continual learning. 

\vfill\pagebreak

\bibliographystyle{IEEEbib}
\bibliography{strings,refs}

\begin{thebibliography}{10}

\bibitem{mirzadeh2020understanding}
S.~I. Mirzadeh, M. Farajtabar, R. Pascanu, and H. Ghasemzadeh,
\newblock ``Understanding the role of training regimes in continual learning,''
\newblock in {\em Proc. Adv. Neural Inf. Process. Syst}, 2020, vol.~33, pp.
  7308--7320.

\bibitem{jerfel2019reconciling}
G. Jerfel, E. Grant, T. Griffiths, and K.~A. Heller,
\newblock ``Reconciling meta-learning and continual learning with online
  mixtures of tasks,''
\newblock in {\em Proc. Adv. Neural Inf. Process. Syst}, 2019, vol.~32, pp.
  9122--9133.

\bibitem{mallya2018packnet}
A. Mallya and S. Lazebnik,
\newblock ``Packnet: Adding multiple tasks to a single network by iterative
  pruning,''
\newblock in {\em Proc. IEEE Comput. Soc. Conf. Comput. Vis. Pattern
  Recognit.}, 2018, pp. 7765--7773.

\bibitem{isele2018selective}
D. Isele and A. Cosgun,
\newblock ``Selective experience replay for lifelong learning,''
\newblock in {\em Proc. AAAI Conf. Artif. Intell.}, 2018, pp. 3302--3309.

\bibitem{rolnick2019experience}
D. Rolnick, A. Ahuja, J. Schwarz, T. Lillicrap, and G. Wayne,
\newblock ``Experience replay for continual learning,''
\newblock in {\em Proc. Adv. Neural Inf. Process. Systs.}, 2019, vol.~32, pp.
  350--360.

\bibitem{kirkpatrick2017overcoming}
J. Kirkpatrick, R. Pascanu, N. Rabinowitz, J. Veness, G. Desjardins, A.~A.
  Rusu, K. Milan, J. Quan, T. Ramalho, A. Grabska-Barwinska, et~al.,
\newblock ``Overcoming catastrophic forgetting in neural networks,''
\newblock in {\em Proc. Nat. Acad. Sci.}, 2017, vol. 114, pp. 3521--3526.

\bibitem{lopez2017gradient}
D. Lopez-Paz and M. Ranzato,
\newblock ``Gradient episodic memory for continual learning,''
\newblock in {\em Proc. Adv. Neural Inf. Process. Systs.}, 2017, vol.~30, pp.
  6467--6476.

\bibitem{saha2021gradient}
G. Saha, I. Garg, and K. Roy,
\newblock ``Gradient projection memory for continual learning,''
\newblock in {\em Proc. Int. Conf. Learn. Represent.}, 2021.

\bibitem{wang2021training}
S. Wang, X. Li, J. Sun, and Z. Xu,
\newblock ``Training networks in null space of feature covariance for continual
  learning,''
\newblock in {\em Proc. IEEE Comput. Soc. Conf. Comput. Vis. Pattern
  Recognit.}, 2021, pp. 184--193.

\bibitem{chaudhry2018efficient}
A. Chaudhry, M. Ranzato, M. Rohrbach, and M. Elhoseiny,
\newblock ``Efficient lifelong learning with a-{GEM},''
\newblock in {\em Proc. Int. Conf. Learn. Represent.}, 2019.

\bibitem{kong2022balancing}
Y. Kong, L. Liu, Z. Wang, and D. Tao,
\newblock ``Balancing stability and plasticity through advanced null space in
  continual learning,''
\newblock in {\em Proc. Eur. Conf. Comput. Vis.}, 2022, pp. 219--236.

\bibitem{liu2018rethinking}
Z. Liu, M. Sun, T. Zhou, G. Huang, and T. Darrell,
\newblock ``Rethinking the value of network pruning,''
\newblock in {\em Proc. Int. Conf. Learn. Represent.}, 2019.

\bibitem{frankle2018the}
J. Frankle and M. Carbin,
\newblock ``The lottery ticket hypothesis: Finding sparse, trainable neural
  networks,''
\newblock in {\em Proc. Int. Conf. Learn. Represent.}, 2019.

\bibitem{Liu_2017_ICCV}
Z. Liu, J. Li, Z. Shen, G. Huang, S. Yan, and C. Zhang,
\newblock ``Learning efficient convolutional networks through network
  slimming,''
\newblock in {\em Proc. IEEE Int. Conf. Comput. Vis.}, 2017, pp. 2736--2744.

\bibitem{aljundi2018memory}
R. Aljundi, F. Babiloni, M. Elhoseiny, M. Rohrbach, and T. Tuytelaars,
\newblock ``Memory aware synapses: Learning what (not) to forget,''
\newblock in {\em Proc. Eur. Conf. Comput. Vis.}, 2018, pp. 139--154.

\bibitem{guo2020improved}
Y. Guo, M. Liu, T. Yang, and T. Rosing,
\newblock ``Improved schemes for episodic memory-based lifelong learning,''
\newblock in {\em Proc. Adv. Neural Inf. Process. Systs.}, 2020, vol.~33, pp.
  1023--1035.

\bibitem{zeng2019continual}
G. Zeng, Y. Chen, B. Cui, and S. Yu,
\newblock ``Continual learning of context-dependent processing in neural
  networks,''
\newblock {\em Nat. Mach. Intell.}, vol. 1, no. 8, pp. 364--372, 2019.

\end{thebibliography}

\end{document}